%% file: main.tex
\documentclass[letterpage]{article}
\usepackage{uai2019}

\usepackage{hyperref}
\hypersetup{colorlinks=true,linkcolor=blue,citecolor=blue}

\usepackage{natbib}
\usepackage[algoruled,algo2e]{algorithm2e}
\usepackage{tabularx}
\SetKwComment{Comment}{$\triangleright$}{}
\usepackage{graphicx}
\usepackage{multirow}
\usepackage{tabularx}
\usepackage{booktabs}
\usepackage{enumitem}

\usepackage{amssymb}
\input{math_commands.tex}

\usepackage{times}

\title{Augmenting and Tuning Knowledge Graph Embeddings}

\author{
{\bf Robert Bamler$^*$}  \\
Department of Computer Science \\
University of California, Irvine \\
Irvine, CA 92617 \\
\And
{\bf Farnood Salehi\thanks{~~joint first authorship.}} \\
\'Ecole Polytechnique F\'ed\'erale \\
de Lausanne (EPFL), \\
Switzerland \\
\And
{\bf Stephan Mandt}   \\
Department of Computer Science \\
University of California, Irvine \\
Irvine, CA 9261
}

\begin{document}

\maketitle

\begin{abstract}
Knowledge graph embeddings rank among the most successful methods for link prediction in knowledge graphs, i.e., the task of completing an incomplete collection of relational facts.
A downside of these models is their strong sensitivity to model hyperparameters, in particular regularizers, which have to be extensively tuned to reach good performance~\citep{KBK2017}. We propose an efficient method for large scale hyperparameter tuning by interpreting these models in a probabilistic framework. After a model augmentation that introduces per-entity hyperparameters, we use a variational expectation-maximization approach to tune thousands of such hyperparameters with minimal additional cost. Our approach is agnostic to details of the model and results in a new state of the art in link prediction on standard benchmark data.
\end{abstract}

\input{intro}
\input{model}
\input{method}
\input{results}
\input{related}
\input{conclusions}

\newpage

\bibliography{references}
\bibliographystyle{plainnat}

\end{document}


\renewcommand{\theequation}{S\arabic{equation}}
\renewcommand{\thepage}{S\arabic{page}}
\renewcommand{\thetable}{S\arabic{table}}
\renewcommand{\thefigure}{S\arabic{figure}}

\maketitle

\input{appendix}

%% file: math_commands.tex
\usepackage{amsmath,amsfonts,bm}

\def\eqref#1{equation~\ref{#1}}

\def\1{\bm{1}}

\def\rmE{{\mathbf{E}}}

\def\rmR{{\mathbf{R}}}

\def\ermE{{\textnormal{E}}}

\def\ermR{{\textnormal{R}}}

\DeclareMathAlphabet{\mathsfit}{\encodingdefault}{\sfdefault}{m}{sl}
\SetMathAlphabet{\mathsfit}{bold}{\encodingdefault}{\sfdefault}{bx}{n}

\def\sB{{\mathbb{B}}}
\def\sC{{\mathbb{C}}}

\def\sR{{\mathbb{R}}}
\def\sS{{\mathbb{S}}}

\def\sV{{\mathbb{V}}}



%% file: intro.tex
\section{INTRODUCTION}\label{sec:introduction}

In 2012, Google announced that it improved the quality of its search engines significantly by utilizing knowledge graphs \citep{E2012}.
A knowledge graph is a data set of relational facts represented as triplets (\emph{head}, \emph{relation}, \emph{tail}).
The \emph{head} and \emph{tail} symbols represent real-world entities, such as people, objects, or places.
The \emph{relation} describes how the two entities are related to each other, e.g., `$\langle$head$\rangle$ \emph{was founded by} $\langle$tail$\rangle$' or `$\langle$head$\rangle$ \emph{graduated from} $\langle$tail$\rangle$'.

While the number of true relational facts among a large set of entities can be enormous, the amount of data points in empirical knowledge graphs is often rather small.
It is therefore desirable to complete missing facts in a knowledge graph algorithmically based on patterns detected in the data set of known facts~\citep{nickel2016review}.
Such link prediction in relational knowledge graphs has become an important subfield of artificial intelligence~\citep{BUGWY2013,WZFC2014,LLSLZ2015,NRP2016,TWR2016,WL2016,JLHZ2016,SHCG2016,XHMZ2017,SW2017,LUO2018}.

A popular approach to link prediction is to fit an embedding model to the observed facts~\citep{KBK2017,N2017,wang2017knowledge}.
A knowledge graph embedding model represents each entity and each relation by a low-dimensional semantic embedding vector.
Over the past six years, these models have made significant progress on link prediction~\citep{BUGWY2013,BWXJL2015,NRP2016,TWR2016,LUO2018}.
However, \citet{KBK2017} pointed out that these models are highly sensitive to hyperparameters, specifically the regularization strength.
This is not surprising since even large knowledge graphs often contain only few data points per entity (i.e., per embedding vector), and so the regularizer plays an important role.
\citet{KBK2017} showed that a simple baseline model can outperform more modern models when using carefully tuned hyperparameters.

In addition to being highly sensitive to the regularization strength, knowledge graph embedding models also need vastly different regularization strengths for different embedding vectors.
Knowledge graph embedding models are typically trained by minimizing some function $\ell_{hrt}$ of the embedding vectors for each triplet fact $(h,r,t)$ (short or \emph{head}, \emph{relation}, and \emph{tail}) in the training set~$\mathbb S$.
One typically adds a regularizer with some strength $\lambda>0$ as follows,
\begin{align}\label{eq:loss_one_lambda}
  \!\!\!\! \sum_{(h,r,t) \in \sS} \!\!\!\!\!\! \Big[
    \ell_{hrt}(\rmE,\rmR)
    + \frac{\lambda}{p} \!\left({
      {||\ermE_h||}_p^p \!+\! {||\ermE_t||}_p^p \!+\! {||\ermR_r||}_p^p
    }\right) \!\Big].
\end{align}
Here, $\ermE_h$, $\ermE_t$, and~$\ermR_r$ is the embedding for entity~$h$, entity~$t$, and relation~$r$, respectively.
Boldface~$\rmE$ and~$\rmR$ is shorthand for all entity and relation embeddings, respectively, and one typically uses a $p$-norm regularizer with $p\in\{2,3\}$.

\begin{figure}
  \centering
  \includegraphics[width=\columnwidth]{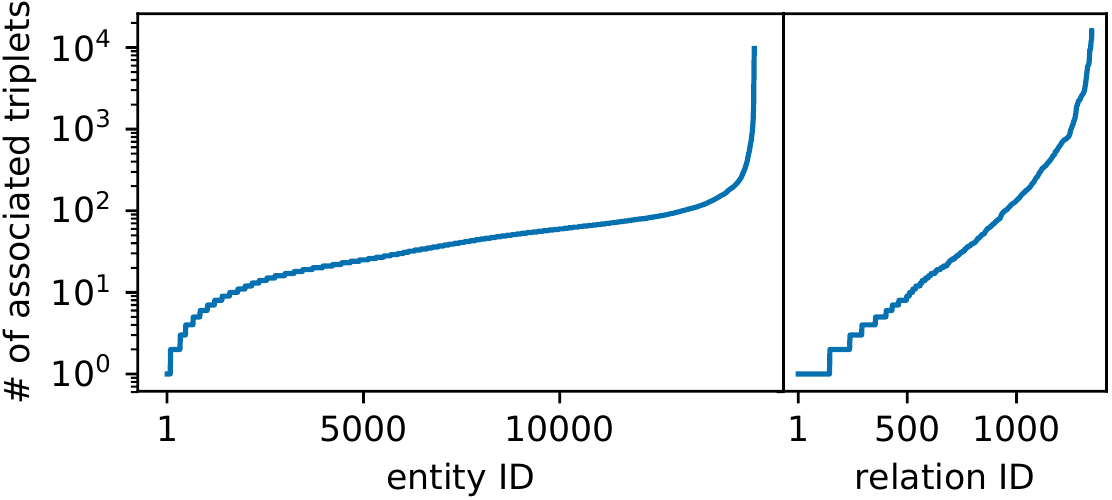}
  \caption{
    Number of training points (triplet facts) for each entity (left) and relation (right) in the FB15K data set (we sorted entity and relation IDs by frequency).
    The large variation on the $y$-axis motivates entity/relation-dependent regularization strengths as proposed in Eq.~\ref{eq:loss_individual_lambdas}.
  }%
  \label{fig:counts}
\end{figure}

It was pointed out by~\citet{LUO2018} that Eq.~\ref{eq:loss_one_lambda} implicitly scales the regularization strength proportionally to the frequency of entities and relations in the data set~$\sS$ since the regularizer is inside the sum over training points.
This implies vastly different regularization strengths for different embedding vectors since the frequencies of entities and relations vary over a wide range (Figure~\ref{fig:counts}).
As we show in this paper, the general idea to use stronger regularization for more frequent entities and relations can be justified from a Bayesian perspective (for empirical evidence, see~\citep{SS2010}).
However, the specific choice to make the regularization strength \emph{proportional} to the frequency seems more like a historic accident.

Rather than imposing a proprotional relationship between frequency and regularization strength, we propose to augment the model family such that each embedding~$\ermE_e$ and~$\ermR_r$ has its individual regularization strength $\lambda^\text{E}_e$ and $\lambda^\text{R}_r$, respectively.
This replaces the loss function from Eq.~\ref{eq:loss_one_lambda} with
\begin{align}\label{eq:loss_individual_lambdas}
  \!\!\!\! \sum_{(h,r,t) \in \sS} \!\!\!\!\! \ell_{hrt}(\rmE, \rmR)
  + \sum_e \frac{\lambda^\text{E}_e}{p} {||\ermE_e||}_p^p
  + \sum_r\frac{\lambda^\text{R}_r}{p} {||\ermR_r||}_p^p.
\end{align}
Here, the last two sums run over each entity~$e$ and each relation~$r$ exactly once (there is only one sum over entities since the same entity embedding vector~$\ermE_e$ is used for an entity~$e$ in either head or tail position).

The loss in Eq.~\ref{eq:loss_individual_lambdas} contains a macroscopic number of hyperparameters $\{\lambda^\text{E}_e\}$~and~$\{\lambda^\text{R}_r\}$:
over $16{,}000$ in our largest experiments.
It would be impossible to tune such a large number of hyperparameters with traditional grid search, which scales exponentially in the number of hyperparameters.
To solve this issue, we propose in this work a probabilistic interpretation of knowledge graph embedding models.
The probabilistic interpretation admits efficient hyperparameter tuning with variational expectation-maximization~\citep{dempster1977maximum,bernardo2003variational}.
This allows us to optimize over all hyperparameters in parallel, and it leads to models with better predictive performance.

Besides improving performance, our approach also has the potential to accelerate research on new knowledge graph embedding models.
Researchers who propose a new model architecture currently have to invest considerable resources into hyperparameter tuning to prove competitiveness with existing, highly tuned models.
Our cheap large-scale hyperparameter tuning speeds up iteration on new models.

In detail, our contributions are as follows:
\begin{itemize}
  \item
    We consider a broad class of knowledge graph embedding models containing ComplEx~\citep{TWR2016} and DistMult~\citep{BWXJL2015}.
    We interpret these models as generative models of facts with a corresponding generative process ( Figure~\ref{fig:generative}).
  \item
    We first \emph{augment} these models by introducing separate priors for each
    entity and relationship vector.
    In a nonprobabilistic picture, these correspond to regularizers.
    This augmentation makes the models more flexibile, but it introduces thousands of new hyperparameters (regularizers) that need to be optimized.
  \item
    We then show how to efficiently \emph{tune} such augmented models.
    The large number of hyperparameters rules out both grid search with cross validation and Bayesian optimization, calling for gradient-based hyperparameter optimization.
    Gradient-based hyperparameter optimization would lead to singular solutions in classical maximum likelihood training.
    Instead, we propose variational expectation-maximization (EM), which avoids such singularities.
  \item
    We evaluate our proposed hyperparameter optimization method experimentally for augmented versions of DistMult and ComplEx.\footnote{Source code: \url{https://github.com/mandt-lab/knowledge-graph-tuning}}
    The high tunability of the proposed models combined with our efficient hyperparameter tuning method improve the predictive performance over the previous state of the art.
\end{itemize}

The paper is structured as follows:
Section~\ref{sec:model} summarizes a large class of knowledge graph embedding models and presents our probabilistic perspective on these models in terms of a generative probabilistic process.
Section~\ref{sec:method} describes our algorithm for hyperparameter tuning.
We present experiments in
Section~\ref{sec:experiments}, compare our method to related work in Section~\ref{sec:related}, and conclude in Section~\ref{sec:conclusions}.

%% file: model.tex
\section{GENERATIVE KNOWLEDGE GRAPH EMBEDDING MODELS}
\label{sec:model}

In this section, we introduce our notation for a large class of knowledge graph embedding models (KG embeddings) from the literature (Section~\ref{sec:conventional_models}), and we then generalize these models in two aspects.
First, while conventional KG embeddings typically share the same regularization strength across all entities and relationship vectors, we lift this constraint and allow each embedding vector to be regularized differently (Section~\ref{sec:regularization}).
Second, we show that the loss functions of conventional KG embeddings as well as our augmented model class can be obtained as point estimates of a probabilistic generative process of the data (Section~\ref{sec:probabilistic_models}).
Drawing on this probabilistic perspective, we can optimize all hyperparameters efficiently using variational expectation-maximization (Section ~\ref{sec:method}).

\subsection{CONVENTIONAL KG EMBEDDINGS}
\label{sec:conventional_models}

We introduce our notation for a large class of knowledge graph embedding models (KG embeddings) from the literature, such as DistMult~\citep{BWXJL2015}, ComplEx~\citep{TWR2016}, and Holographic Embeddings~\citep{NRP2016}.

Knowledge graphs are sets of triplet facts $(h,r,t)$ where the `head'~$h$ and `tail'~$t$ both belong to a fixed set of~$N_\text{e}$ entities, and~$r$ describes which one out of a set of~$N_\text{r}$ relations holds between~$h$ and~$t$.
KG embeddings represent each entity $e\in[N_\text{e}]$ and each relation $r\in[N_\text{r}]$ by an embedding vector~$\ermE_e$ and~$\ermR_r$, respectively, that lives in a semantic embedding space~$\sV$ with a low dimension~$K$.
A model is defined by a real valued score function~$f(\ermE_h,\ermR_r,\ermE_t)$.
One fits the embedding vectors such that~$f$ assigns a high score to observed triplet facts~$(h,r,t)\in\mathbb S$ in the training set~$\mathbb S$ and a low score to triplets that do not appear in~$\mathbb S$.

\paragraph{Examples.}
We give examples of the two models that reach highest predictive performance to the best of our knowledge.
For more models, see~\citep{KBK2017}.

In the DistMult model~\citep{BWXJL2015}, the embedding space $\sV=\sR^K$ is real valued, and the score is defined as
\begin{align}\label{eq:score_distmult}
  f(\ermE_h,\ermR_r,\ermE_t) &= \sum_{k=1}^K \ermE_{hk} \ermR_{rk} \ermE_{tk}
  \qquad\text{(DistMult)}
\end{align}
where, e.g., $\ermE_{hk}\in\sR$ is the $k$\textsuperscript{th} entry of the vector~$\ermE_h$.

The ComplEx model~\citep{TWR2016} uses a complex embedding space $\sV= \sC^K$, and defines the score
\begin{align}\label{eq:score_complex}
  f(\ermE_h,\ermR_r,\ermE_t) &= \sum_{k=1}^K \text{Re}\!\left[
    \ermE_{hk} \ermR_{rk} \bar\ermE_{tk}
  \right]
  \quad\text{(ComplEx)}
\end{align}
where $\text{Re}[\,\cdot\,]$ denotes the real part of a complex number, and~$\bar\ermE_{tk}$ is the complex conjugate of~$\ermE_{tk}\in\sC$.

\paragraph{Tail And Head Prediction.}
Typical benchmark tasks for KG embeddings are `tail prediction' and `head prediction', i.e., completing queries of the form $(h,r,?)$ and $(?,r,t)$, respectively, by ranking potential completions $(h,r,t)$ by their score $f(\ermE_h,\ermR_r,\ermE_t)$.
Most proposals for KG embeddings train a single model for both tail and head prediction.
Thus, the loss function is given by Eq.~\ref{eq:loss_one_lambda}, where $\ell_{hrt}(\rmE,\rmR)$ is a sum of two terms to train for tail and head prediction, respectively.
While early works (e.g.,~\citep{BUGWY2013,WZFC2014,BWXJL2015}) trained by maximizing a margin over negative samples, the more recent literature~\citep{KBK2017,LKHJ2018} suggests that the softmax loss leads to better predictive performance,

\begin{align} \label{eq:individual_loss}
  \ell_{hrt}(\rmE,\rmR)
  = &- f(\ermE_h,\ermR_r,\ermE_t) + \log\Bigg(\sum_{t'=1}^{N_\text{e}} e^{f(\ermE_h,\ermR_r,\ermE_{t'\!})}\Bigg) \nonumber\\
  & - f(\ermE_h,\ermR_r,\ermE_t) + \log\Bigg(\sum_{h'=1}^{N_\text{e}} e^{f(\ermE_{h'\!},\ermR_r,\ermE_t)}\Bigg).
\end{align}
Here, the first line (with the sum over tails~$t'$) is the softmax loss for tail prediction, while the second line (with the sum over heads~$h'$) is the softmax loss for head prediction.

\subsection{REGULARIZATION IN KG EMBEDDINGS}
\label{sec:regularization}

Knowledge graph embedding models are highly sensitive to hyperparameters, especially to the strength of the regularizer~\citep{KBK2017}.
This can be understood since even large knowledge graphs typically contain only few data points per entity.
For example, the FB15K data set contains $483{,}000$ data points, but $88\%$ of all entities $e\in[N_\text{e}]$ appear fewer than $100$ times as head or tail of a training point.
Moreover, the amount of training data varies strongly across entities and relations (see Figure~\ref{fig:counts}), suggesting that the regularization strength for embedding vectors~$\ermE_e$ and~$\ermR_r$ should depend on the entity~$e$ and relation~$r$.

The loss function for conventional KG embeddings in Eq.~\ref{eq:loss_one_lambda} regularizes all embedding vectors with the same strength~$\lambda$.
We propose to replace~$\lambda$ by individual regularization strengths~$\lambda^\text{E}_e$ and~$\lambda^\text{R}_r$ for each entity~$e$ and relation~$r$, respectively, and to fit models with the loss function in Eq.~\ref{eq:loss_individual_lambdas}.
It generalize Eq.~\ref{eq:loss_one_lambda}, which one obtains for
\begin{align} \label{eq:proportional_lambdas}
    \lambda^\text{E}_e = \lambda n^\text{E}_e;
    \quad \lambda^\text{R}_r = \lambda n^\text{R}_r;
    \quad\text{(conventional models)}
\end{align}
where $n^\text{E}_e$ and $n^\text{R}_r$ denote the number of times that entity~$e$ or relation~$r$ appears in the training data, respectively.
The proposed augmented models described by Eq.~\ref{eq:loss_individual_lambdas} are more flexible as they do not impose a linear relationship between~$n^\text{E/R}_{e/r}$ and the regularization strength~$\lambda^\text{E/R}_{e/r}$.

The downside of the augmented KG embedding models is that one has to tune a macroscopic number of hyperparameters $\{\lambda^\text{E}_e\}$ and~$\{\lambda^\text{R}_r\}$:
more than $16{,}000$ in the popular FB15K data set.
Tuning such a large number of hyperparameters would be far too computationally expensive in a conventional setup that fits point estimates by minimizing the loss function.
For point estimated models, it is well known that one cannot fit hyperparameters to the training data as this would lead to overfitting (see also Supplementary Material).
To avoid overfitting, knowledge graph embedding models are conventionally tuned by cross validation on heldout data.
This requires training a model from scratch for each new hyperparameter setting.
Cross validation does not scale beyond models with a handful of hyperparameters, and it is expensive even there (see, e.g.,~\citep{KBK2017,LUO2018}).

Probabilistic models, by contrast, allow tuning of many hyperparameters in parallel using the empirical Bayes method~\citep{dempster1977maximum,maritz2018empirical}.
We propose a probabilistic formulation of augmented KG embeddings in the next section, and we present a method for efficient hyperparameter tuning in these models in Section~\ref{sec:method}.

\subsection{PROBABILISTIC KG EMBEDDINGS}
\label{sec:probabilistic_models}

We now present our probabilistic version of KG embeddings.
The probabilistic formulation enables efficient optimization over thousands of hyperparameters, see Section~\ref{sec:method}.

\begin{figure}
  \centering
  \includegraphics[width=\columnwidth]{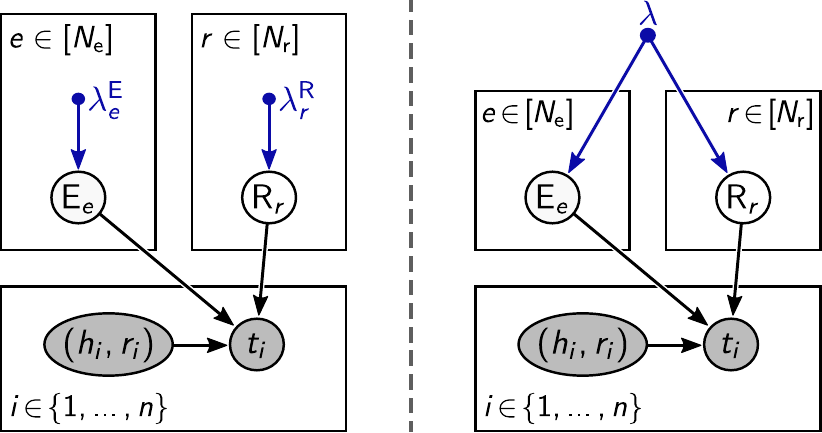}
  \caption{Two different generative processes of triplet facts ${\{(h_i,r_i,t_i)\}}_{i=1}^n$.
  Hyperparameters (regularizer strengths) highlighted in blue.
  Left: proposed class of models augmented with individual regularizer strengths~$\lambda^\text{E/R}_{e/r}$.
  Right: probabilistic interpretation of conventional models~(Eq.~\ref{eq:loss_one_lambda}).
  The conventional models fix the relative strength of regularizers by reducing all~$\lambda^\text{E/R}_{e/r}$ to a single scalar~$\lambda$.}%
  \label{fig:generative}
\end{figure}

\paragraph{Reciprocal Facts.}
The KG embedding models discussed in Sections~\ref{sec:conventional_models} and~\ref{sec:regularization} make a direct interpretation as a generative probabilistic process difficult.
Training a single model for both head and tail prediction introduces cyclic causal dependencies.
As will become clear below, the tail prediction part in Eq.~\ref{eq:individual_loss} (first line on the right-hand side) corresponds to a generative process where the head~$h$ causes the tail~$t$.
However, the head prediction part (second line) corresponds to a generative process where~$t$ causes~$h$.

To solve this issue, we employ a data augmentation due to~\citet{LUO2018} that goes as follows.
For each relation $r\in [N_\text{r}]$, one introduces a new symbol~$r^{-1}$, which has the interpretation of the inverse of~$r$, but whose embedding vector is not tied to that of~$r$.
One then constructs an augmented training set~$\sS'$ by adding the reciprocal facts,
\begin{align}\label{eq:augmented_dataset}
  \sS' := \sS \cup {\{(t, r^{-1}, h)\}}_{(h,r,t) \in \sS}
\end{align}
One trains the model by minimizing the loss in Eq.~\ref{eq:loss_one_lambda} or Eq.~\ref{eq:loss_individual_lambdas}, where the sum over data points is now over~$\sS'$ instead of~$\sS$, and where $\ell_{hrt}$ is given by only the first line of Eq.~\ref{eq:individual_loss}.
When evaluating the model performance on a test set, one answers head prediction queries $(?, r, t)$ by answering the corresponding tail prediction query $(t, r^{-1}, ?)$.
This data augmentation was introduced in~\citep{LUO2018} to improve performance.
As we show next, it also has the advantage of enabling a probabilistic interpretation by establishing a causal order where~$h$ comes before~$t$.

\paragraph{Generative Process.}
With the above data augmentation, minimizing the loss function in Eq.~\ref{eq:loss_individual_lambdas} is equivalent to point estimating the parameters of the probabilistic graphical model shown in Figure~\ref{fig:generative} (left).
The generative process is:
\begin{itemize}
  \item
    For each entity $e\in[N_\text{e}]$ and each relation $r\in[N_\text{r}]$, draw an embedding $\ermE_e, \ermR_r\in\sV$ from the priors
    \begin{equation}\label{eq:priors}
    \begin{aligned}
      p(\ermE_e|\lambda^\text{E}_e) \propto e^{-\frac{\lambda^\text{E}_e}{p} {||\ermE_e||}_p^p}\,; \quad
      p(\ermR_r|\lambda^\text{R}_r) \propto e^{-\frac{\lambda^\text{R}_r}{p} {||\ermR_r||}_p^p}\,.
    \end{aligned}
    \end{equation}
    Here, $p\in\{2,3\}$ specifies the norm, and the omitted proportionality constant follows from normalization.
  \item
    Repeat for each data point to be generated:
    \begin{itemize}[topsep=-1pt,leftmargin=15pt]
      \item
        Draw a head entity $h$ and a relation $r$ from some discrete distribution $P(h, r)$.
        The choice of this distribution has no influence on inference since~$h$ and~$r$ are both directly observed.
      \item
        Draw a tail entity
        \begin{align} \label{eq:categorical}
            t \sim \text{Categorical}\!\left( \text{softmax}_t(f(\ermE_h,\ermR_r,\ermE_t) \right)
        \end{align}
        where $f$ is the score (e.g., Eq.~\ref{eq:score_distmult} or~\ref{eq:score_complex}), and
        \begin{align} \nonumber
          \text{softmax}_t(f(\ermE_h,\ermR_r,\ermE_t)) &= \frac{e^{f(\ermE_h,\ermR_r,\ermE_t)}}{\sum_{t'} e^{f(\ermE_h,\ermR_r,\ermE_{t'})}}.
        \end{align}
      \item
        Add the triplet fact $(h, r, t)$ to the data set~$\sS'$.
    \end{itemize}
\end{itemize}

This process defines a log joint distribution over~$\rmE$,~$\rmR$, and the data $\sS'$, conditioned on the hyperparameters~$\{\lambda^\text{E/R}_{e/r}\}$, which we denote collectively by the boldface symbol~$\boldsymbol\lambda$,
\begin{align}
  &\log p(\rmE, \rmR, \sS'| \boldsymbol\lambda) = \nonumber\\
  &\qquad = \!\!\sum_{e\in [N_\text{e}]}\!\! \log p(\ermE_e | \lambda^\text{E}_e)
     +\!\!\sum_{r\in [N_\text{r}]}\!\! \log p(\ermR_r | \lambda^\text{R}_r) \nonumber\\
    &\qquad +\!\!\!\!\!\sum_{(h,r,t)\in\sS'}\!\!\!\!\! \big[\!
        \log P(h, r) + \log p(t | h, r, \rmE, \rmR) \big]. \label{eq:log_joint}
\end{align}
Using Eq.~\ref{eq:categorical}, it is easy to see that $\log p(t | h, r, \rmE, \rmR)$ is the negative of the first line of Eq.~\ref{eq:individual_loss}.
Thus, up to an additive term that depends only on~$\boldsymbol\lambda$, the log joint distribution in Eq.~\ref{eq:log_joint} is the negative of the loss function, and minimizing the loss over~$\rmE$ and~$\rmR$ is equivalent to a maximum a posteriori (MAP) approximation of the probabilistic model.

Figure~\ref{fig:generative} compares the generative process of the augmented KG embeddings proposed in Section~\ref{sec:regularization} (left part of the figure) to the generative process for conventional KG embeddings that one obtains by setting~$\lambda^\text{E}_e$ and~$\lambda^\text{R}_r$ as in Eq.~\ref{eq:proportional_lambdas} (right).
The augmented models are more flexible due to the large number of hyperparameters~$\lambda^\text{E/R}_{e/r}$.
We discuss next how the probabilistic interpretation allows us to efficiently optimize over this large number of hyperparameters.

%% file: method.tex
\section{HYPERPARAMETER OPTIMIZATION}
\label{sec:method}

We now describe the proposed method for hyperparameter tuning in the probabilistic knowledge graph embedding models introduced in Section~\ref{sec:probabilistic_models}.
The method is based on variational expectation-maximization (EM).
We first derive an approximate coordinate update equation for the hyperparameters (Section~\ref{sec:variational_em}) and then cover details of the parameter initialization (Section~\ref{sec:algorithm_phases}).

Variational EM optimizes a lower bound to the marginal likelihood of the model over hyperparameters $\boldsymbol\lambda$, with model parameters $\rmE$ and $\rmR$ integrated out.
As we show in the supplementary material, the naive alternative of simultaneously optimizing the original model's loss function
over model parameters and hyperparameters would lead to divergent solutions.
Variational EM avoids such divergent solutions by keeping track of parameter uncertainty.
We elaborate on the role of parameter uncertainty in the supplementary material.

\begin{algorithm2e}[t!]
  \setlength{\hsize}{\dimexpr(\columnwidth-0.8em)}
  \SetKwComment{Comment}{$\triangleright\;$}{}
  \DontPrintSemicolon
  \SetKwFor{Repeat}{repeat}{times}{end}
  \SetKwRepeat{Repeatuntil}{repeat}{until}
  \begin{tabularx}{\hsize}{@{}l@{$\,\,$}>{\raggedright\arraybackslash}X@{}}
    \textbf{Input:}
      & Augmented data set $\sS'$, see Eq.~\ref{eq:augmented_dataset}. \\
      & Model $p(\rmE, \rmR, \sS' | \boldsymbol\lambda)$ as defined by Eqs.~\ref{eq:priors}-\ref{eq:log_joint}.\\
      & Hyperparameters $\boldsymbol\lambda$.
        Learning rate $\alpha$. \\[2pt]
    \textbf{Output:} & Trained embedding vectors $\rmE$ and $\rmR$.
  \end{tabularx}\\\vspace{2pt}
  \nl Initialize embedding vectors $\rmE$ and $\rmR$ randomly.\;
  \nl \Repeatuntil{\rm convergence (see Section~\ref{sec:algorithm_phases})}{
    \nl Draw a minibatch $\sB \subset \sS'$.\;
    \nl Calculate a minibatch estimate $\hat L_\sB$ of the loss $L(\rmE,\rmR|\boldsymbol\lambda) := -\log p(\rmE, \rmR, \sS' | \boldsymbol\lambda)$.\;
    \nl Update $\rmE \gets \rmE - \alpha \nabla_{\!\rmE} \hat L_\sB$ and $\rmR \gets \rmR - \alpha \nabla_{\!\rmR} \hat L_\sB$.
  }
  \caption{Conventional training of knowledge graph embedding models (stochastic gradient descent).}
  \label{alg:sgd}
\end{algorithm2e}

\subsection{VARIATIONAL EM FOR KNOWLEDGE GRAPH EMBEDDING MODELS}
\label{sec:variational_em}

Our proposed algorithm based on variational EM can easily be implemented in an existing model architecture by making a few modifications.
Algorithm~\ref{alg:sgd} shows the conventional way to train a knowledge graph embedding model using stochastic gradient descent (SGD).
The log joint distribution in Eqs.~\ref{eq:priors}-\ref{eq:log_joint} defines a loss function $L(\rmE,\rmR|\boldsymbol\lambda) := -\log p(\rmE,\rmR,\sS'|\boldsymbol\lambda)$ of the form of Eq.~\ref{eq:loss_individual_lambdas}.
In SGD, one repeatedly calculates an estimate~$\hat L_\sB$ of this loss function based on a minibatch~$\sB$ of training points, and one obtains gradient estimates by backpropagating through~$\hat L_\sB$.

Algorithm~\ref{alg:variational_em} shows the modifications that are necessary to implement hyperparameter optimization.
We describe the algorithm in detail below.
In summary, one has to:
(i)~inject noise into the loss estimate~$\hat L_\sB$ (lines~\ref{ln:draw_noise}-\ref{ln:inject_noise});
(ii)~learn the optimal amount~$\boldsymbol\xi$ of noise via SGD (line~\ref{ln:update_xi}); and
(iii)~update the hyperparameters~$\boldsymbol\lambda$ (line~\ref{ln:update_lambda}).

\begin{algorithm2e}[t!]
  \setlength{\hsize}{\dimexpr(\columnwidth-0.8em)}
  \SetKwComment{Comment}{$\triangleright\;$}{}
  \DontPrintSemicolon
  \SetKwFor{Repeat}{repeat}{times}{end}
  \SetKwRepeat{Repeatuntil}{repeat}{until}
  \begin{tabularx}{\hsize}{@{}l@{$\,\,$}>{\raggedright\arraybackslash}X@{}}
    \textbf{Input:}
      & Augmented data set $\sS'$, see Eq.~\ref{eq:augmented_dataset}. \\
      & Model $p(\rmE, \rmR, \sS' | \boldsymbol\lambda)$ as defined by Eqs.~\ref{eq:priors}-\ref{eq:log_joint}.\\
      & Initial values for $\boldsymbol\mu$, $\boldsymbol\xi$, $\boldsymbol\lambda$, see Section~\ref{sec:algorithm_phases}. \\
      & Numbers $T_\text{E}$, $T_\text{EM}$ of E-steps and EM-steps. \\
      & Learning rates $\alpha_{\boldsymbol\mu}, \alpha_{\boldsymbol\xi} >0$ and $\alpha_{\boldsymbol\lambda}\in (0,1]$. \\[2pt]
    \textbf{Output:} & Optimized hyperparameters $\boldsymbol\lambda$, embedding vectors $\boldsymbol\mu^\text{E/R} \pm e^{\boldsymbol\xi^\text{E/R}}$ with uncertainties.
  \end{tabularx}\\\vspace{2pt}
  \nl Initialize means~$\boldsymbol\mu^\text{E/R}$ and log standard deviations~$\boldsymbol\xi^\text{E/R}$ around a pretrained model, see Section~\ref{sec:algorithm_phases}.\;\vspace{2pt}
  \nl \For{$t\leftarrow 1$ \KwTo $T_{\rm E} + T_{\rm EM}$}{
    \nl Draw a minibatch $\sB \subset \sS'$.\;
    \nl Draw Gaussian noise $\boldsymbol\epsilon^\text{E}, \boldsymbol\epsilon^\text{R} \sim \mathcal N(0, I)$. \label{ln:draw_noise}\;
    \nl Calculate a minibatch estimate $\hat L_\sB$ of the loss with injected noise, $L(\boldsymbol\mu^\text{E} + e^{\boldsymbol\xi^\text{E}} \!\odot \boldsymbol\epsilon^\text{E}, \boldsymbol\mu^\text{R} + e^{\boldsymbol\xi^\text{R}} \!\odot \boldsymbol\epsilon^\text{R} | \boldsymbol\lambda)$ (``$\odot$'' denotes elementwise multiplication). \label{ln:inject_noise}\;
    \nl Update $\boldsymbol\mu \gets \boldsymbol\mu - \alpha_{\boldsymbol\mu} \nabla_{\!\boldsymbol\mu} \hat L_\sB$.\;
    \nl Update $\boldsymbol\xi \gets \boldsymbol\xi - \alpha_{\boldsymbol\xi} (\nabla_{\!\boldsymbol\xi} \hat L_\sB - \mathbf{1})$. \label{ln:update_xi}\;
    \vspace{3pt}
    \nl \If{$t > T_{\rm E}$}{
      \nl Update $\boldsymbol\lambda \gets \big[(1 - \alpha_{\boldsymbol\lambda}) \boldsymbol\lambda^{-1} + \alpha_{\boldsymbol\lambda} \boldsymbol{\hat\lambda}^{-1} \big]^{-1}$. \label{ln:update_lambda}\;
      \Comment*[r]{\rm\textit{M-step (see Eq.~\ref{eq:optimal_lambda}).}}
      \vspace{-2.67ex}
    }
  }
  \caption{Variational EM for knowledge graph embedding models with coordinate updates for $\boldsymbol\lambda$.}
  \label{alg:variational_em}
\end{algorithm2e}

\paragraph{Variational Expectation-Maximization.}
Our probabilistic interpretation of knowledge graph embedding models allows us to optimization over all hyperparameters $\{\lambda^\text{E}_e\}$ and $\{\lambda^\text{R}_r\}$ in parallel via the expectation-maximization (EM) algorithm~\citep{dempster1977maximum}.
This algorithm treats the model parameters~$\rmE$ and~$\rmR$ as latent variables that have to be integrated out.
The EM algorithm alternates between a step in which the latent variables are integrated out (`E-step'), and an update step for the hyperparameters~$\boldsymbol\lambda$ (`M-step').
We use a version of EM based on variational inference, termed variational EM \citep{bernardo2003variational}, that avoids the integration step.
We further derive an approximate coordinate update equation for the hyperparameters~$\boldsymbol\lambda$, which lead to a significant speedup over gradient updates in our experiments.

Each choice of hyperparameters~$\boldsymbol\lambda$ defines a different variant of the model.
The marginal likelihood of the data,
\begin{align} \label{eq:marginal_likelihood}
  p(\sS'|\boldsymbol\lambda) &= \int p(\rmE,\rmR, \sS' |\boldsymbol\lambda)\, d\rmE \,d\rmR
\end{align}
quantifies how well a given model variant describes the data~$\sS'$.
Maximizing $p(\sS' | \boldsymbol\lambda)$ over $\boldsymbol\lambda$ thus yields the model variant that fits the data best.
However, $p(\sS' | \boldsymbol\lambda)$ is unavailable in closed form as the integral in Eq.~\ref{eq:marginal_likelihood} is intractable.

To circumvent the problem of the intractable marginal likelihood, we use variational inference (VI) \citep{JGJS1999}.
Rather than integrating over the entire space of model parameters $\rmE$ and $\rmR$, we maximize a lower bound on the marginal likelihood.
We introduce a so-called variational family of Gaussian probability distributions,
\begin{align} \label{eq:q1}
  q_{\boldsymbol\mu,\boldsymbol\sigma}(\rmE, \rmR)
  &= q_{\boldsymbol\mu^\text{E},\boldsymbol\sigma^\text{E}}(\rmE)\;
     q_{\boldsymbol\mu^\text{R},\boldsymbol\sigma^\text{R}}(\rmR)
\end{align}
with
\begin{align} \label{eq:q2}
  q_{\boldsymbol\mu^\text{E},\boldsymbol\sigma^\text{E}}(\rmE)
  &= \prod_{e\in[N_\text{e}]} \prod_{k=1}^K \mathcal N(\ermE_{ek}; \mu_{ek}, \sigma_{ek}^2)
\end{align}
and analogously for $q_{\boldsymbol\mu^\text{R},\boldsymbol\sigma^\text{R}}(\rmR)$.
Here, the means $\boldsymbol\mu\equiv(\boldsymbol\mu^\text{E},\boldsymbol\mu^\text{R})$ and the standard deviations $\boldsymbol\sigma\equiv(\boldsymbol\sigma^\text{E},\boldsymbol\sigma^\text{R})$ are so-called variational parameters over which we optimize.

Evoking Jensen's inequality, the log marginal likelihood is then lower-bounded by the \emph{evidence lower bound} \citep{BKM2017,ZBKM2017}, or ELBO:
\begin{align}
  &\log p(\sS'|\boldsymbol\lambda) \nonumber\\
  &\quad \geq \mathbb{E}_{\rmE,\rmR\sim q_{\boldsymbol\mu,\boldsymbol\sigma}} \big[
    \log p(\rmE,\rmR, \sS' | \boldsymbol\lambda)
    - \log q_{\boldsymbol\mu,\boldsymbol\sigma}(\rmE,\rmR) \big] \nonumber\\
  &\quad = -\mathbb{E}_{\rmE,\rmR\sim q_{\boldsymbol\mu,\boldsymbol\sigma}} \big[
    L(\rmE,\rmR, \sS' | \boldsymbol\lambda)]
    + H[q_{\boldsymbol\mu,\boldsymbol\sigma} ] \nonumber\\
  &\quad =: \text{ELBO}(\boldsymbol\lambda, \boldsymbol\mu,\boldsymbol\sigma). \label{eq:elbo}
\end{align}
Here, in the second step, we identified the log joint probability as the negative of the loss $L$ of the corresponding point estimated model, and $H[q_{\boldsymbol\mu,\boldsymbol\sigma}]$ is the entropy of $q_{\boldsymbol\mu,\boldsymbol\sigma}$.

The bound in Eq.~\ref{eq:elbo} is tight if the variational distribution~$q_{\boldsymbol\mu,\boldsymbol\sigma}$ is the true posterior of the model for given $\boldsymbol\lambda$.
Since it is a lower bound, maximizing the ELBO over~$\boldsymbol\mu$ and $\boldsymbol\sigma$ minimizes the gap and yields the best approximation of the marginal likelihood.
We thus take the ELBO as a proxy for the marginal likelihood, and we maximize it also over $\boldsymbol\lambda$ to find near-optimal hyperparameters.

\paragraph{Gradient updates for $\boldsymbol\mu$ and $\boldsymbol\sigma$.}
We maximize the ELBO concurrently over both variational parameters $\boldsymbol\mu$ and $\boldsymbol\sigma$ as well as over hyperparameters $\boldsymbol\lambda$.
Updating the variational parameters is called the ``E-step''.
Here, we use gradient updates using Black Box reparameterization gradients \citep{KW2014,RMW2014}.
This has the advantage of being agnostic to the model architecture as long as the score $f(\ermE_h,\ermR_r,\ermE_t)$ (e.g., Eqs.~\ref{eq:score_distmult}-\ref{eq:score_complex}) is differentiable, and it requires only few changes compared to the standard SGD training loop in Algorithm~\ref{alg:sgd}.

To make sure that the standard deviations are always positive, we parameterize them by their logarithms~$\boldsymbol\xi$,
\begin{align}
  \boldsymbol\sigma = e^{\boldsymbol\xi}
  \qquad\text{(elementwise)},
\end{align}
and we optimize over~$\boldsymbol\mu$ and~$\boldsymbol\xi$ using SGD.
We obtain an unbiased estimate of the term $\mathbb{E}_{q_{\boldsymbol\mu,\boldsymbol\sigma}}[L]$ in Eq.~\ref{eq:elbo} by drawing a single sample from $q_{\boldsymbol\mu,\boldsymbol\sigma}$ (lines~\ref{ln:draw_noise}-\ref{ln:inject_noise} in Algorithm~\ref{alg:variational_em}).
The reparameterization gradient trick uses the fact that for,~e.g., noise $\epsilon^\text{E}_{ek} \sim \mathcal N(0,1)$ from a standard normal distribution, $\mu^\text{E}_{ek} + \sigma^\text{E}_{ek} \epsilon^\text{E}_{ek}$ is distributed as $\mathcal N\big(\mu^\text{E}_{ek}, {(\sigma^\text{E}_{ek})}^2\big)$.
The entropy part $H[q_{\boldsymbol\mu,\boldsymbol\sigma}]$ of the ELBO (Eq.~\ref{eq:elbo}) can be calculated analytically.
Up to an additive constant, it is given by the sum over all log standard deviations $\xi^\text{E}_{ek}$~and~$\xi^\text{R}_{rk}$.
Thus, its gradient with respect to $\boldsymbol\xi$ has the constant value of one in each coordinate direction, which we denote by the bold face term~``$\boldsymbol{1}$'' on line~\ref{ln:update_xi} of Algorithm~\ref{alg:variational_em}.

\paragraph{Coordinate updates for $\boldsymbol\lambda$.}
Optimizing the ELBO over $\boldsymbol\lambda$ leads to an improved set of hyperparameters provided that the ELBO is a good approximation of the marginal likelihood $p(\sS' | \boldsymbol\lambda)$.
However, this is typically not the case at the beginning of the optimization when the variational distribution is still a poor fit of the posterior.
We therefore begin the optimization with some number $T_\text{E}$ of pure ``E-step'' updates during which we keep $\boldsymbol\lambda$ fixed.
After $T_\text{E}$ ``E-steps'', we alternate between ``E'' and ``M'' steps, where the latter update the hyperparameters~$\boldsymbol\lambda$.
In our experiments, we found that the optimization converged slowly when we used gradient updates for~$\boldsymbol\lambda$.
To speed up convergence, we therefore derive approximate coordinate updates for $\boldsymbol{\lambda}$.

To simplify the notation, we derive the update equation only for a single hyperparameter~$\lambda^\text{E}_e$.
Updates for~$\lambda^\text{R}_r$ are analogous.
The only term in the ELBO (Eq.~\ref{eq:elbo}) that depends on $\lambda^\text{E}_e$ is the expected log prior, $\mathbb{E}_{q_{\boldsymbol\mu,\boldsymbol\sigma}}[\log p(\ermE_e|\lambda^\text{E}_e)]$.
Since this term is independent of the data we can write it out explicitly.
The omitted proportionality constant in the prior (Eq.~\ref{eq:priors}) is dictated by normalization.
We find,
\begin{align}\label{eq:prior_normalization}
  \log p(\ermE_e | \lambda^\text{E}_e) =
    \frac{K'}{p} \log\big(\lambda^\text{E}_e\big) - \frac{\lambda^\text{E}_e}{p} {||\ermE_e||}_p^p
    + \text{const,}
\end{align}
where $K'=K$ for a real-valued embedding space $\sV = \sR^K$ (as in DistMult) and $K'=2K$ if $\sV = \sC^K$ (as in ComplEx).
Setting $\nabla_{\!\lambda^\text{E}_e} \mathbb{E}_{q_{\boldsymbol\mu,\boldsymbol\sigma}}[\log p(\ermE_e|\lambda^\text{E}_e)]$ to zero we find that the regularizer strength $\hat\lambda^\text{E}_e$ that maximizes the ELBO for given $\boldsymbol\mu$ and $\boldsymbol\sigma$ satisfies
\begin{align} \label{eq:optimal_lambda}
  \frac{1}{\hat\lambda^\text{E}_e}
  = \frac{1}{K'}\, \mathbb{E}_{q_{\boldsymbol\mu,\boldsymbol\sigma}}\big[{||\ermE_e||}_p^p\big]
\end{align}

In moderately high embedding dimensions~$K$, we can approximate the right-hand side of Eq.~\ref{eq:optimal_lambda} accurately by sampling from $q_{\boldsymbol\mu,\boldsymbol\sigma}$.
It is the expectation of the average of a large number of independent random variables, and therefore follows a highly peaked distribution.
The update step on line~\ref{ln:update_lambda} of Algorithm~\ref{alg:variational_em} uses a conservative weighted average between the current and the optimal value of $1/\lambda^\text{E}_e$ with a learning rate $\alpha_{\boldsymbol\lambda} \in (0, 1]$.
This effectively averages the estimates over past training steps with a decaying weight.
Note that, for $\sV=\sR^K$, Eq.~\ref{eq:optimal_lambda} has a closed form solution for $p=2$ and $p=3$, but we found it unnecessary in our experiments to implement specialized code for these cases.

\paragraph{Absence of overfitting.}
While the variational EM algorithm keeps track of uncertainty of model parameters, it fits only point estimates for the hyperparameters~$\boldsymbol\lambda$.
This is justified in our setup since there are much fewer hyperparameters than model parameters:
each entity~$e$ and each relation~$r$ has an embedding vector~$\ermE_e$ or~$\ermR_r$ with $K=2{,}000$ scalar components in our experiments, but only a single scalar hyperparameter~$\lambda^\text{E}_{e}$ or~$\lambda^\text{E}_r$.
We therefore expect much smaller posterior uncertainty for~$\boldsymbol\lambda$ than for~$\rmE$ and~$\rmR$, which justifies point estimating~$\boldsymbol\lambda$.
Had we instead chosen a very flexible prior distribution with many hyperparameters per entity and relation, the EM algorithm would have essentially fitted the prior to the variational distribution, leading to an ill-posed problem.
Judging from learning curves on the validation set, we did not detect any overfitting in variational EM.

\subsection{PRE- AND RE-TRAINING}
\label{sec:algorithm_phases}

Variational EM (Algorithm~\ref{alg:variational_em}) converges more slowly than fitting point estimates (Algorithm~\ref{alg:sgd}) because the injected noise increases the variance of the gradient estimator.
To speed up convergence, we train the model in three consecutive phases: pre-training, variational EM, and re-training.

In the pre-training phase, we keep the hyperparameters~$\boldsymbol\lambda$ fixed and fit point estimates $\rmE$~and~$\rmR$ to the model using standard SGD (Algorithm~\ref{alg:sgd}).
We found the final predictive performance (after the variational EM and re-training phases) to be insensitive to the initial hyperparameters.
We use early stopping based on the mean reciprocal rank (see Eq.~\ref{eq:mrr} in Section~\ref{sec:experiments} below), evaluated on the validation set.

In the variational EM phase (Algorithm~\ref{alg:variational_em} and Section~\ref{sec:variational_em}), we initialize the variational distribution $q_{\boldsymbol\mu,\boldsymbol\sigma}$ around the pre-trained model parameters $\rmE$ and $\rmR$.
In detail, we initialize $\boldsymbol\mu^\text{E} \gets \rmE$ and $\boldsymbol\mu^\text{R} \gets \rmR$, and we initialize the components of $\boldsymbol\sigma$ with a value that is small compared to the typical components of $\boldsymbol\mu$ (0.2 in our experiments).

In the re-training phase, we fit again point estimates~$\rmE$ and~$\rmR$ with Algorithm~\ref{alg:sgd}, this time using the optimized hyperparameters~$\boldsymbol\lambda$.
We use the resulting models to evaluate the predictive performance, see results in Section~\ref{sec:experiments}.

Alternatively to re-training a point estimated model, one could also perform predictions by averaging predictive probabilities over samples from the variational distribution~$q_{\boldsymbol\mu,\boldsymbol\sigma}$.
If $q_{\boldsymbol\mu,\boldsymbol\sigma}$ is a good approximation of the model posterior then this results in an approximate Bayesian form of link prediction.
In our experiments, we found that, in low embedding dimensions~$K \lesssim 100$, predictions based on samples from $q_{\boldsymbol\mu,\boldsymbol\sigma}$ outperformed predictions based on point estimates.
In higher embedding dimensions, however, the point estimated models from the re-training phase had better predictive performance.
We interpret this somewhat counterintuitive observation as a failure of the fully factorized Gaussian variational approximation to adequately approximate the true posterior.

%% file: results.tex
\begin{table*}[tb]
  \caption{
  Model performances (in `filtered' setting; see Eqs.~\ref{eq:mrr}-\ref{eq:hits_at_10}).
  $^\ast$\citet{LUO2018} report performance metrics only with two decimals.
  In order to show three decimals, we reproduced their results using the code provided by the authors.
  When rounding to two digits, we recover all values reported in~\citep{LUO2018} except that the MRR for \hbox{FB15K-237} is reported there as 0.37.
  The small discrepancy in the last decimal may be explained by different random seeds.
  }
  \centering
  \begin{tabularx}{\textwidth}{llr@{$\;\;$}c@{$\;\;$}cc@{$\;\;$}cc@{$\;\;$}cc@{$\;\;$}c}
    \toprule
    \multicolumn{3}{r@{$\;\;$}}{data set $\rightarrow$}
      & \multicolumn{2}{c}{WN18RR}
      & \multicolumn{2}{c}{WN18}
      & \multicolumn{2}{c}{FB15K-237}
      & \multicolumn{2}{c}{FB15K} \\
    $\downarrow$ model & $\downarrow$ variant & metric $\rightarrow$
      & MRR & Hits@10
      & MRR & Hits@10
      & MRR & Hits@10
      & MRR & Hits@10 \\\midrule\midrule
    DistMult & \multicolumn{2}{l@{$\quad$}}{\citet{BWXJL2015} (orig.)}
      & -- & --
      & 0.83 & 0.942
      & -- & --
      & 0.35 & 0.577 \\
    DistMult & \multicolumn{2}{l@{$\quad$}}{\citet{KBK2017}}
      & -- & --
      & 0.790 & 0.950
      & -- & --
      & 0.837 & 0.904 \\
    DistMult & \multicolumn{2}{l@{$\quad$}}{\citet{DMSR2017}}
      & 0.43 & 0.49
      & 0.822 & 0.936
      & 0.241 & 0.419
      & 0.654 & 0.824 \\
    DistMult & \multicolumn{2}{l@{$\quad$}}{Ours (after variational EM)}
      & \textbf{0.455} & \textbf{0.544}
      & \textbf{0.911} & \textbf{0.961}
      & \textbf{0.357} & \textbf{0.548}
      & \textbf{0.841} & \textbf{0.914} \\\midrule
    ComplEx & \multicolumn{2}{l@{$\quad$}}{\citet{TWR2016} (orig.)}
      & -- & --
      & 0.941 & 0.947
      & -- & --
      & 0.692 & 0.840 \\
    ComplEx & \multicolumn{2}{l@{$\quad$}}{\citet{LUO2018}$^\ast$}
      & 0.478 & 0.569
      & 0.952 & 0.963
      & 0.364  & 0.555
      & \textbf{0.857} & 0.909 \\
    ComplEx & \multicolumn{2}{l@{$\quad\;\;\;$}}{Ours (after variational EM)}
      & \textbf{0.486} & \textbf{0.579}
      & \textbf{0.953} & \textbf{0.964}
      & \textbf{0.365}& \textbf{0.560}
      & 0.854 & \textbf{0.915} \\\bottomrule
  \end{tabularx}\label{table:results}
\end{table*}

\section{EXPERIMENTAL RESULTS}
\label{sec:experiments}

We test the performance of the proposed model augmentation and the scalable hyperparameter tuning algorithm with two models and four different data sets.
In this section, we report results using standard benchmark metrics and we compare to the previous state of the art.
We also analyze the relationship between the optimized regularizer strengths and the frequency of entities and relations.

\paragraph{Model architectures and baselines.}
We report results for the DistMult model~\citep{BWXJL2015} and the ComplEx model~\citep{TWR2016}.
We follow \citep{LUO2018} for details of the model architecture: we use reciprocal facts as described at the end of Section~\ref{sec:probabilistic_models}, $(p=3)$-norm regularizers, and an embedding dimension of $K=2000$.
We compare our results to the previous state of the art:
\citep{DMSR2017,KBK2017} for DistMult and \citep{LUO2018} for ComplEx.

\paragraph{Data sets.}
We used four standard data sets.
The first two are FB15K from the Freebase project~\citep{BEPST2008} and WN18 from the WordNet database~\citep{BGWB2014}.
The other two data sets, FB15K-237 and WN18RR, are modified versions of FB15K and WN18 due to~\citep{TC2015,DMSR2017}.
The motivation for the modified data sets is that FB15K and WN18 contain near duplicate relations that lead to leakage into the test set, which makes link prediction trivial for some facts, thus encouraging overfitting.
In FB15K-237 and WN18RR these near duplicates were removed.

\paragraph{Metrics.}
We report two standard metrics used in the KG embedding literature: mean reciprocal rank (MRR) and Hits@10.
We average over head and tail prediction on the test set~$\sS_\text{test}$, which is equivalent to averaging only over tail prediction on the augmented test set~$\sS_\text{test}'$, see Eq.~\ref{eq:augmented_dataset}.

All results are obtained in the `filtered' setting introduced in~\citep{BUGWY2013}, which takes into account that more than one tail may be a correct answer to a query $(h,r,?)$.
When calculating the rank of the target tail~$t$ one therefore ignores any competing tails~$t'$ if the corresponding fact $(h,r,t')$ exists in either the training, validation, or test set.
More formally, the fitlered rank, denoted below as $\text{rank}_\text{filt.}(t|h,r)$, is defined as one plus the number of `incorrect' facts $(h,r,t')$ with the the given~$h$ and~$r$ for which $f(\ermE_h,\ermR_r,\ermE_{t'}) \geq f(\ermE_h,\ermR_r,\ermE_t)$.
Here, candidate facts $(h,r,t')$ are considered `incorrect' if they appear neither in the training nor in the validation or test set.

Given a test set~$\sS_\text{test}'$ with reciprocal facts (Eq.~\ref{eq:augmented_dataset}), the mean reciprocal rank (MRR) is (see, e.g., \citep{BWXJL2015}),
\begin{align} \label{eq:mrr}
  \text{MRR} = \frac{1}{|\sS_\text{test}'|}
    \sum_{(h,r,t)\in \sS_\text{test}'}\frac{1}{\text{rank}_\text{filt.}(t|h,r)}.
\end{align}
With `Hits@10', we denote the fraction of test queries for which the filtered rank is at most 10,
\begin{align} \label{eq:hits_at_10}
  \text{Hits@10} = \frac{|\{(h,r,t)\in \sS_\text{test}': \text{rank}_\text{filt.}(t|h,r) \leq 10\}|}{|\sS_\text{test}'|}.
\end{align}

\paragraph{Quantitative results.}
Table~\ref{table:results} summarizes our quantitative results.
The top half of the table shows results for the DistMult model.
Our models with individually optimized regularization strengths significantly outperform the previous state of the art across all four data sets.

For the ComplEx model, the performance improvements are less pronounced (lower half of Table~\ref{table:results}).
This may be explained by the fact that the results in~\citep{LUO2018} were already obtained after large scale expensive hyperparameter tuning using grid search.
By contrast, the hyperparameter search with our proposed method required only a single run per data set.
Even for the largest data set FB15K, the variational EM phase took less than three hours on a single GPU.
Despite the much cheaper hyperparameter optimization, our models slightly outperform the previous state of the art on three out of the four considered data sets, with only a small degradation on the fourth.

\begin{figure}[t!]
  \centering
  \includegraphics[width=\columnwidth]{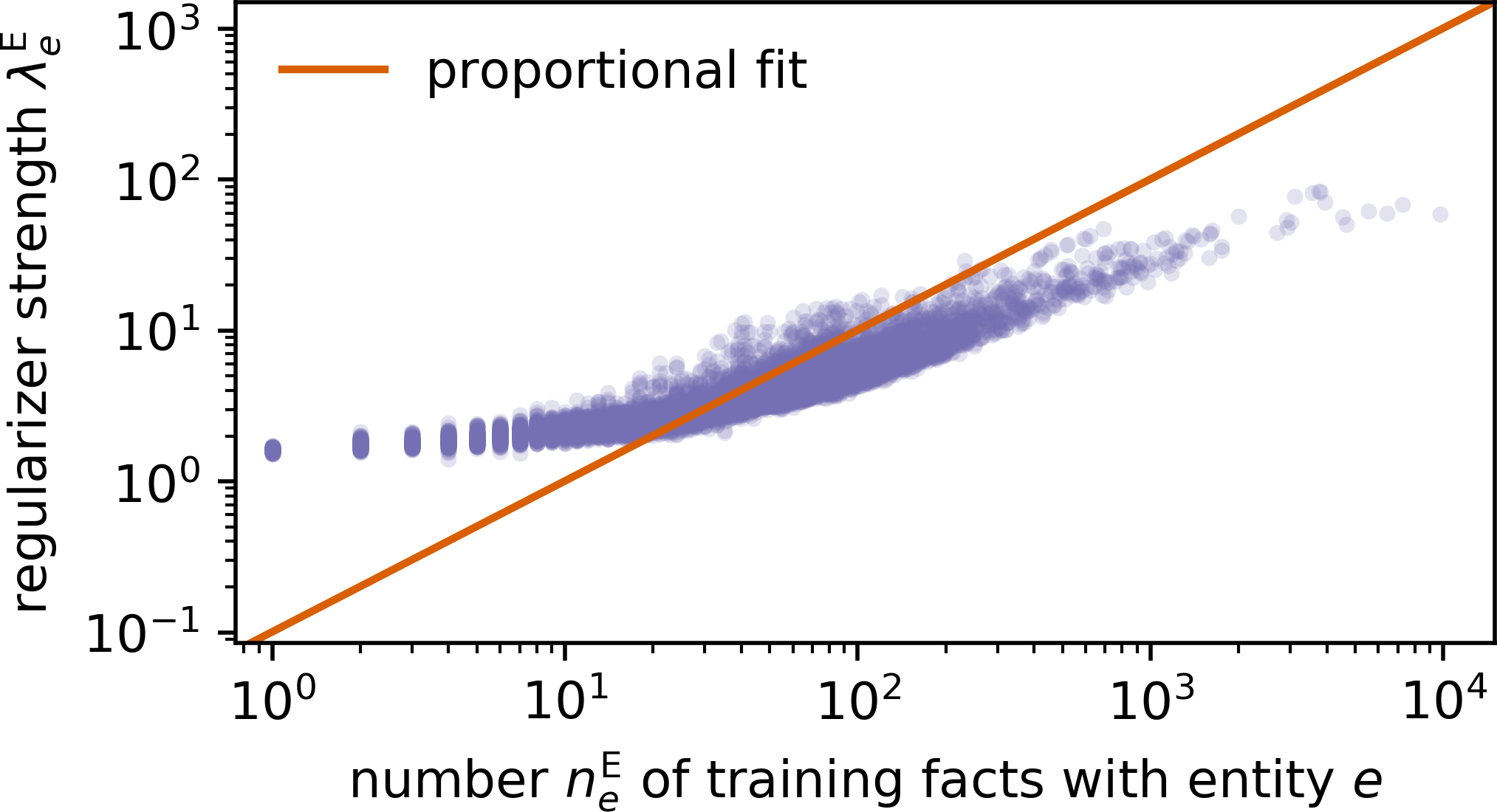}
  \caption{Relationship between the learned regularizer strengths $\lambda^\text{E}_e$ and the frequency of each entity.
  The conventional proportional scaling (red line) overregularizes frequent entities and underregularizes infrequent ones.}\label{fig:lambda_freq}
\end{figure}

\paragraph{Qualitative results.}
Finally, we study the relationship between optimized hyperparameters and frequencies of entities in the training data.
Figure~\ref{fig:lambda_freq} shows the learned $\lambda^\text{E}_e$ for all entities $e$ as a function of the number of times $n^\text{R}_e$ that each entity $e$ appears in the training corpus of the FB15K data set.
The red line is the best proportional fit $\lambda^\text{E}_e \propto n^\text{E}_e$ to the results, as is imposed by conventional models (Eq.~\ref{eq:proportional_lambdas}).

Our findings confirm the general idea to use stronger regularization for entities with more training data.
The Bayesian interpretation can explain this observation:
a small amount of training data typically leads to high posterior uncertainty, which leads to small $\hat\lambda^\text{E}_e$ in Eq.~\ref{eq:optimal_lambda}.
However, our results indicate that imposing a proportionality between $\lambda^\text{E}_e$ and $n^\text{E}_e$ would be a poor choice that significantly underregularizes infrequent entities and overregularizes frequent entities (note the logarithmic scale in Figure~\ref{fig:lambda_freq}).
Our empirical findings may inspire future theoretical work that derives an optimal frequency dependency of the regularization strength in tensor factorization models.

%% file: related.tex
\section{RELATED WORK}\label{sec:related}
Related work to this paper can be grouped into link prediction algorithms and variational inference.

\paragraph{Link Prediction.}
Link prediction in knowledge graphs has gained a lot of attention as it may guide a way towards automated reasoning with real world data.
For a review see \citep{nickel2016review}.
Two different approaches for link prediction are predominant in the literature.
In statistical relational learning, one infers explicit rules about relations (such as transitivity or commutativity) by detecting statistical patterns in the training set.
One then uses these rules for logic reasoning~\citep{FGKP1999,KNP2011,NZRS2012,PMGC2015}.

Our work focuses on a complementary approach that builds on knowledge graph embedding models.
This line of research started with the proposal of the TransE model~\citep{BUGWY2013}, which models relational facts as vector additions in a semantic space.
More recently, a plethora of different knowledge graph embedding models based on tensor factorizations have been proposed.
We summarize here only the path that lead to the current state of the art.
Different models make different trade-offs between generality and effective use of training data.

Canonical tensor decomposition~\citep{H1927} uses independent embeddings for entities in the head or tail position of a fact.
DistMult~\citep{BWXJL2015,TC2015}, by contrast, uses the same embeddings for entities in head and tail position, thus making use of more training data per entity embedding, but restricting the model to symmetric relations.
The ComplEx model~\citep{TWR2016} lifts this restriction by multiplying the head, relation, and tail embeddings in an asymmetric way.
To the best of our knowledge, the current state of the art was presented by \citet{LUO2018}, who improved upon the ComplEx model by introducing reciprocal relations and using a better regularizer.

The sensitivity of KG embeddings to the choice of hyperparameters, such as regularizer strengths, was first pointed out in~\citep{KBK2017}.
A popular heuristic is to regularize each embedding every time it appears in a minibatch, thus effecitvely regularizing embeddings proportionally to their frequency~\citep{SS2010,LUO2018}.
In contrast, we propose to learn entity-dependent regularization strengths without relying on heuristics.

\cite{vilnis2018probabilistic} proposed a new model that is probabilistic in the sense that it assigns probabilities to the results of queries.
However, in contrast to our proposal, the model is not a probabilistic generative model of the data set.

\paragraph{Variational Inference.} Variational inference (VI) is a powerful technique to approximate a Bayesian posterior over latent variables given observations \citep{JGJS1999,BKM2017,ZBKM2017}. Besides approximating the posterior, VI also estimates the marginal likelihood of the data. This allows for iterative hyperparameter tuning, (variational EM) \citep{bernardo2003variational}, which is the main benefit of the Bayesian approach used in this paper.

Our paper builds on recent probabilistic extensions of embedding models to Bayesian models, such as word \citep{B2017} or paragraph \citep{JBSM2017} embeddings. In these works, the words are embedded into a $K$-dimensional space. It has been shown that using a probabilistic approach leads to better performance on small data sets, and allows these models to be combined with powerful priors, such as for time series modeling~\citep{BM2017,bamler2018improving,JWKM2018}. Yet, the underlying probabilistic models in these papers are very different from the ones considered in our work.

\paragraph{Bayesian Optimization.}
An alternative method that can be used for hyperparameter optimization is Bayesian optimization.
However, Bayesian optimization does not scale to the large number of hyperparameters that we tune in this work.
Most practical applications of Bayesian optimization (e.g.,~\citep{snoek2012practical,wang2013bayesian}) tune only tens of hyperparameters, rather than ten thousands.
This is because Bayesian optimization treats the model as a black box, which it can only train and then evaluate for a given choice of hyperparameters at a time.
Each such evaluation contributes a single data point to fit an auxiliary model over the hyperparameters.
By contrast, variational EM has access to gradient information to train all hyperparameters in parallel, and concurrently with the model parameters.

%% file: conclusions.tex
\section{CONCLUSIONS}
\label{sec:conclusions}

We augmented a large class of popular knowledge graph embedding models in such a way that every entity embedding and every relationship embedding vector has their own regularizer, and showed that it is possible to tune these potentially thousands of hyperparameters in a scalable way. Our approach is motivated by the observation that sharing a common regularization parameter across all embeddings leads to over-regularization.

We treat knowledge graph embeddings as generative probabilistic models, making them amenable to Bayesian model selection.
We derived approximate coordinate updates for the hyperparameters in the framework of variational EM.
We applied our method to generalizations of the DistMult and ComplEx models and outperformed the state of the art for link prediction. The approach
can be applied to a wide range of models with minimal modifications to the training routine. In the future, it would be interesting to investigate whether tighter variational bounds~\citep{burda2016importance,bamler2017perturbative} may further improve model selection.

%% file: appendix.tex
\section*{The Role of Parameter Uncertainty in the Proposed Hyperparameter Optimization}

Bayesian inference and the idea of measuring uncertainty are somewhat uncommon in the literature for knowledge graph embeddings.
To clarify why uncertainty is important in the proposed hyperparameter optimization (Section~3.1 of the main text), we compare the method here to a more naive approach that will turn out to fail because it ignores uncertainty.
We discuss the failure of the naive approach and the benefit of estimating parameter uncertainty first intuitively and then more formally.

\paragraph{Intuitive picture.}
The variational EM algorithm maximizes (a lower bound on) the marginal likelihood $p(\sS'|\boldsymbol\lambda)$, Eq.~14 of the main text.
In a more naive attempt to hyperparameter tuning without cross validation, one might be tempted to skip the marginalization over model parameters $\rmE$ and $\rmR$.
Instead, one might try to directly maximize the log joint probability $\log p(\rmE, \rmR, \sS' | \boldsymbol\lambda)$, i.e., minimize the loss $L=-\log p(\rmE, \rmR, \sS' | \boldsymbol\lambda)$, over $\rmE$, $\rmR$, and hyperparameters $\boldsymbol\lambda$.
This approach, however, would lead to divergent solutions because the log joint probability is unbounded.

\begin{figure}[t!]
  \centering
  \includegraphics[width=\columnwidth]{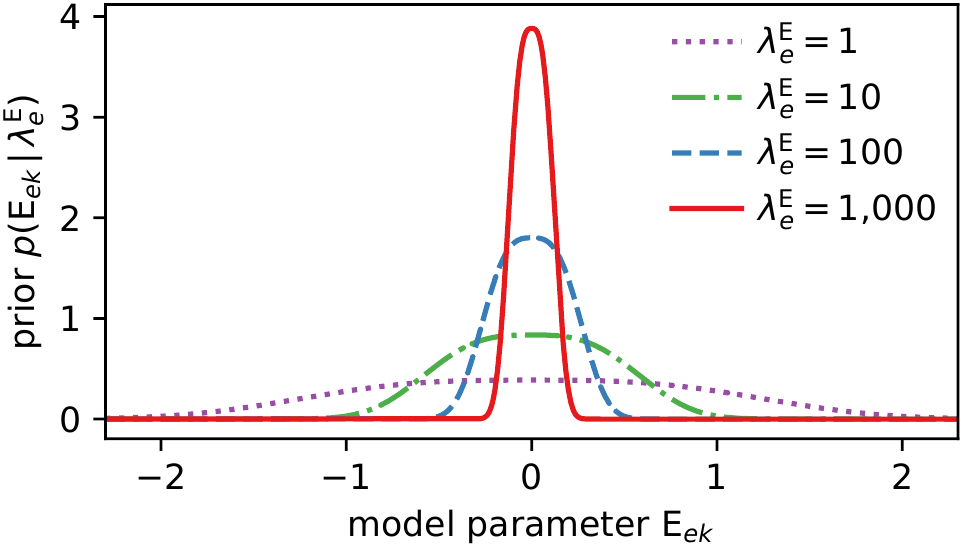}
  \caption{Prior (Eq.~8 of the main text with $p=3$) for a single model parameter~$\ermE_{ek}$.
  As the prior gets more peaked for growing regularizer strength~$\lambda^\text{E}_e$, the height of the peak grows unboundedly.
  An (incorrect) hyperparameter optimization method that ignores parameter uncertainty could end up exploiting this unboundedness and diverge to~$\lambda^\text{E}_e \to \infty$.
  }\label{fig:priors}
\end{figure}

The log joint probability contains the log priors (first two terms on the right-hand side of Eq.~10 of the main text).
Figure~\ref{fig:priors} shows the prior $p(\ermE_{ek}|\lambda^\text{E}_e)$, Eq.~8 of the main text, of a single component $\ermE_{ek}$ of an entity embedding assuming, for simplicity, a real embedding space.
With growing regularizer strength (increasing~$\lambda^\text{E}_e$), the prior becomes narrower and narrower.
As the peak narrows, it also grows higher due to the normalization constraint.
In the limit $\lambda^\text{E}_e \to\infty$, the prior collapses to an infinitely narrow and high $\delta$-peak at zero.

A hyperparameter tuning method that ignores posterior uncertainty can exploit the unbounded growth of the maximum of the prior to send the log joint distribution to infinity (i.e., the loss $L\to-\infty$).
Without any posterior uncertainty, one can set the model parameter $\ermE_{ek}$ precisely to zero and then make the value of $p(\ermE_{ek}|\lambda^\text{E}_e)$ arbitrarily large by sending $\lambda^\text{E}_{ek}\to\infty$.

We prevent this collapse of the prior to a $\delta$-peak by keeping track of parameter uncertainty.
Admitting a nonzero uncertainty for $\ermE_{ek}$ no longer allows us to set~$\ermE_{ek}$ precisely and deterministically to zero.
Any slightly nonzero value of~$\ermE_{ek}$ would have no support under a $\delta$-peaked prior.

\paragraph{Formal derivation.}
We now formalize the above intuitive picture and show that the specific variational approximation chosen in Eqs.~12-13 indeed suffices to prevent any divergent solutions.

Assuming again a real embedding space, the log prior of a single model parameter~$\ermE_{ek}$ is given by (cf., Eq.~16 of the main text)
\begin{align} \label{eq-supp:logprior}
  \log p(\ermE_e | \lambda^\text{E}_e) &=
    \frac{1}{p}\Big[\! \log\lambda^\text{E}_e - \lambda^\text{E}_e {|\ermE_{ek}|}^p \Big]
    + \text{const.}
\end{align}
As discussed above, setting $\ermE_{ek}=0$ and sending $\lambda^\text{E}_e\to\infty$ sends the right-hand side of Eq.~\ref{eq-supp:logprior} to infinity.
This can even be relaxed:
the log prior diverges for $\lambda^\text{E}_e\to\infty$ as long as we keep $\ermE_{ek}$ small enough, i.e., as long as $\ermE_{ek} = O((\lambda^\text{E}_e)^{-1/p})$.
This is why maximizing the log joint distribution over~$\boldsymbol\lambda$ leads to divergend solutions.

Instead of maximizing the log joint distribution, the variational EM algorithm maximizes the ELBO, Eq.~14 of the main text.
Note first that the ELBO itself is bounded from above by zero:
the ELBO is a lower bound on the marginal log likelihood $\log p(\sS'|\boldsymbol\lambda)$, where $p(\sS'|\boldsymbol\lambda)\leq 1$ since it is a discrete probability distribution.

Further, the ELBO has a maximum at finite values for the variational parameters and hyperparameters.
Maximizing the ELBO ensures that each model parameter is associated with a nonzero uncertainty $\sigma^\text{E/R}_{e/r}>0$ since the entropy term
\begin{align} \label{eq-supp:entropy}
    H[q_{\boldsymbol\mu,\boldsymbol\sigma}]
    &= \sum_{e\in [N_\text{e}]} \log\sigma^\text{E}_e
      +\sum_{r\in [N_\text{r}]} \log\sigma^\text{R}_r + \text{const.}
\end{align}
imposes an infinite penalty if any $\sigma^\text{E/R}_{e/r} \to0$.
The entropy term in the ELBO thus has an additional regularizing effect.
Combined with the other regularizing term in the ELBO, the expected log prior, we obtain for a given model parameter~$\ermE_{ek}$ using Eq.~\ref{eq-supp:logprior}, up to an additive constant,
\begin{align}
    &\mathbb E_{q_{\boldsymbol\mu,\boldsymbol\sigma}}\left[
        \log p(\ermE_{ek} | \lambda^\text{E}_e)
        -\log \big(q_{\sigma^\text{E}_{ek},\mu^\text{E}_{ek}}\!(\ermE_{ek})\big)
        \right] \nonumber\\
    &\;\; =
        \frac{1}{p} \log\lambda^\text{E}_e
        - \frac{\lambda^\text{E}_e}{p}\mathbb{E}_{q_{\boldsymbol\sigma,\boldsymbol\mu}}\big[
            {\left| \ermE_{ek} \right|}^p \big]
        + \log\sigma^\text{E}_{ek} \label{eq-supp:regterms1}\\
    &\;\; =\frac{1}{p}\!\left[
        \log\left(\lambda^\text{E}_e {\left(\sigma^\text{E}_{ek}\right)}^p \right)
        - \lambda^\text{E}_e\,
          \mathbb{E}_{\epsilon\sim\mathcal N(0,1)}\!\left[
            {\left| \mu^\text{E}_{ek} + \sigma^\text{E}_{ek} \epsilon \right|}^p \right]
    \right] \nonumber
\end{align}
where, in the last equality, we made the dependency on $\sigma^\text{E}_{ek}$ explict by reparameterizing the normal distributed random variable~$\ermE_{ek} = \mu^\text{E}_{ek} + \sigma^\text{E}_{ek}\epsilon$ in terms of a standard-normal distributed variable~$\epsilon$.

Maximizing the right-hand side of Eq.~\ref{eq-supp:regterms1} over~$\mu^\text{E}_{ek}$ yields $\mu^\text{E}_{ek}=0$ by symmetry, thus simplifying the objective to
\begin{align} \label{eq-supp:regterms2}
    &\frac{1}{p}\!\left[
        \log\left(\lambda^\text{E}_e {\left(\sigma^\text{E}_{ek}\right)}^p \right)
        - \lambda^\text{E}_e {\left(\sigma^\text{E}_{ek}\right)}^p
            \;\mathbb{E}_{\epsilon\sim\mathcal N(0,1)}\big[{\left|\epsilon \right|}^p \big]
    \right] \nonumber\\
    =& \frac{1}{p}\!\left[
        \log\left(\lambda^\text{E}_e {\left(\sigma^\text{E}_{ek}\right)}^p \right)
        - \lambda^\text{E}_e {\left(\sigma^\text{E}_{ek}\right)}^p \,c_p
    \right]
\end{align}
with some numerical constant $c_p>0$.
The right-hand side of Eq.~\ref{eq-supp:regterms2} is structurally similar to the right-hand side of Eq.~\ref{eq-supp:logprior}, which had divergent solutions:
both are a difference between a logarithmic and a linear term in~$\lambda^\text{E}_e$.
However, while in Eq.~\ref{eq-supp:logprior}, we were able to send the logarithmic term to infinity and still keep the linear term bounded, this is not possible in Eq.~\ref{eq-supp:regterms2}, which has the same argument (up to a constant~$c_p$) for the logarithmic and the linear term.
The right-hand side of Eq.~\ref{eq-supp:regterms2} has a maximum at a finite value for $\lambda^\text{E}_e {\left(\sigma^\text{E}_{ek}\right)}^p$.
Thus, the variational EM algorithm avoids divergent solutions.